\documentclass{Interspeech}

\title{Linguistically Informed Tokenization Improves ASR for Underresourced Languages}

\author[affiliation={1}]{Massimo}{Daul}
\author[affiliation={2}]{Alessio}{Tosolini}
\author[affiliation={2}]{Claire}{Bowern}

\affiliation{Department of Mathematics}{New York University}{USA}
\affiliation{Department of Linguistics}{Yale University}{USA}
\email{mmd9604@nyu.edu, alessio.tosolini@yale.edu, claire.bowern@yale.edu}

\usepackage{comment}
\usepackage{subcaption}
\newcommand{\ngg}{\textipa{N}} % requires \usepackage{tipa}

\begin{document}

\maketitle

\begin{abstract}
    
Automatic speech recognition (ASR) is a crucial tool for linguists aiming to perform a variety of language documentation tasks. However, modern ASR systems use data-hungry transformer architectures, rendering them generally unusable for underresourced languages. We fine-tune a wav2vec2 ASR model on Yan-nhangu, a dormant Indigenous Australian language, comparing the effects of phonemic and orthographic tokenization strategies on performance. In parallel, we explore ASR's viability as a tool in a language documentation pipeline. We find that a linguistically informed phonemic tokenization system substantially improves WER and CER compared to a baseline orthographic tokenization scheme. Finally, we show that hand-correcting the output of an ASR model is much faster than hand-transcribing audio from scratch, demonstrating that ASR can work for underresourced languages.
    
\end{abstract}

\section{Introduction}
% Thinks to look into: \\
% - \cite{Diwan2020ReduceAR} Reduce and Reconstruct: ASR for Low-Resource Phonetic Languages \\
% - \cite{9687921} Comparative Study of Different Tokenization Strategies for Streaming End-to-End ASR \\
% - \cite{10094667} Massively Multilingual ASR on 70 Languages: Tokenization, Architecture, and Generalization Capabilities \\
% - \cite{khare2021low} Low Resource ASR: The Surprising Effectiveness of High Resource Transliteration. \\
% - \cite{yeroyan2024enablingasrlowresourcelanguages} Enabling ASR for Low-Resource Languages: A Comprehensive Dataset Creation Approach \\
% - \cite{bekarystankyzy_multilingual_2024} Multilingual end-to-end ASR for low-resource Turkic languages with common alphabets \\
% - \cite{atuhurra2024} Introducing Syllable Tokenization for Low-resource Languages: A Case Study with Swahili \\
% - \todo{Add stuff about tokenization or how people think about sounds, stuff with phonemes}

% suggest also ELPis as example for low resource languages: https://www.isca-archive.org/interspeech_2019/foley19_interspeech.pdf or https://research.google/pubs/managing-transcription-data-for-automatic-speech-recognition-with-elpis/
% Outline of the paragraphs in the introduction:
Automatic Speech Recognition (ASR, also known as speech-to-text) is a natural language processing technology that converts spoken words to text. Most modern ASR systems are neural and transformer-based. For example, wav2vec2 \cite{baevski2020wav2vec20frameworkselfsupervised} uses self-supervised learning to train ASR models on large amounts of unlabeled speech data, while Whisper \cite{radford2022robustspeechrecognitionlargescale} is an encoder-decoder model trained on diverse multilingual and multitask speech data. ASR is very widely used for a broad range of human--computer interaction tasks, as well as accessibility services such as captioning and subtitling, and in many types of research where searchable text needs to be generated from speech. There is a big resource gap, however, where only a small number of languages have well performing and freely available ASR models. Common Voice \cite{ardila2020commonvoicemassivelymultilingualspeech}, for example, covers only 2\% of the world's languages. The geographic distribution of these languages is also unequal, with no Indigenous Australian languages, such as Yan-nhangu, represented in this corpus.  

One reason for this asymmetry is that where the majority of speech--text training data comes from linguistic fieldwork, manual transcription and annotation is a major bottleneck. It is time-consuming and requires specialist knowledge, and few documentation projects have the resources to train and pay annotators \cite{chelliah2001}. Transcription can take anywhere from 5 minutes to over an hour per minute of speech \cite{dwyer2006} and documentation projects often generate more recordings than there is time to transcribe. That means that there is both much less training data for under-resourced languages, and the means to acquire such training data is extremely labor intensive. Accurate ASR would vastly assist in this respect. 
%\textcolor{orange}{In linguistic research, transcription and annotation is a major bottleneck}. \todo{Add information here - Claire}

ASR models train by predicting a sequence of tokens, where each token corresponds to a piece of text. During training, features are extracted from the audio and passed to the model. The model generates a sequence of token state probabilities in a forward pass, which is then converted into a joint-probability maximizing token sequence, which is the predicted transcription. This sequence is compared to the labeled transcription, and the model adjusts its parameters to try to minimize a predetermined loss function. 
% For optimal performance, ASR models benefit from many hours of clean, diarized, and non-overlapping audio, as this facilitates the learning of relationships between audio features and tokens. 
Tokenization is the first step in processing or generating any type of continuous textual data. In the context of ASR, tokenization refers to the process of segmenting transcribed speech into meaningful units, such as words, subwords, or characters. Different tokenization strategies impact vocabulary size (the smaller and more general each token, the smaller the total vocabulary size), error rates, and adaptability to various languages and domains. Byte Pair Encoding (BPE) and its variants are the most commonly used type of textual tokenization which work by continuously expanding the vocabulary through the creation of new, larger tokens from combining smaller tokens. However, for certain low-resource neural tasks including ASR, performance is increased through smaller more finegrained tokens \cite{adlaon-marcos-2024-finding, banerasroux:hal-04584931}. BPE struggles with accurately encoding linguistic information \cite{bostrom2020bytepairencodingsuboptimal, Limisiewicz_2024}, with linguistically informed tokenization strategies resulting in improved model performance for some low-resource tasks \cite{atuhurra2024}. This paper investigates whether a linguistically informed phonemic tokenization improves ASR model accuracy for Yan-nhangu compared to a baseline orthographic tokenization, exploring the impact of linguistic knowledge on low-resource NLP systems.

%There are several ways in which writing systems encode phonological information for spoken languages. Many writing systems are phonemic (at the syllabic or segmental level); others are logographic. Segmental scripts may encode segmental and supersegmental information, such as tone; or may omit segments (as abjads, which do not represent most vowels). Orthographies may use a single segment per phoneme (in a 1:1 mapping), or be under- or over-deterministic. 
There is much variation in the relationship between the orthography of a language and the phonological information it encodes. For most practical applications, the final output of an ASR model is preferred to be orthographic, even though it may not be the best representative of the underlying phonological information encoded in the audio. Many languages, such as the one at the focus of this paper, have a phonology fully predictable from the orthography. Although a transparent mapping is common across the world's languages, it is important to note that there are many ways that the orthography and phonology of a language can be related to and derived from one another.  

\section{Methodology}
\subsection{Data origin and preprocessing}
The corpus for this test includes selected recordings of the Yan-nhangu language (Glottolog code \textsc{yann1237}; ISO-639 \textsc{jay}). Recordings were made from 5 fluent speakers of Yan-nhangu between 2004 and 2007. Four of the five were from Nha\textipa{N}u clans; the fifth was Warrawarra. The traditional lands of Yan-nhangu people are the Crocodile Islands of Northeast Arnhem Land (Northern Territory, Australia). Yan-nhangu is a member of the Yol\textipa{N}u subgroup of Pama-Nyungan. All five were women and recognised by their community as authorities to speak about the language. At the time of recording they ranged in age from early 40s to early 70s. %A sixth, very senior authority for Yan-nhangu also worked closely with the language documentation project.

Recordings were made with an Edirol R-01 solid state recorder and external microphone. They were recorded in a range of locations in Milingimbi Aboriginal Community and on Murru\textipa{N}ga Island. Language tasks included a range of structured and semi-structured elicitation tasks, including wordlist and sentence translation, descriptions of pictures and video clips, and discussion prompts around cultural concepts and practices. Materials were transcribed in Elan \cite{wittenburgELANProfessionalFramework2006a} by the last author and checked with speakers. Recordings, transcripts, and field notes have been deposited with the ELAR digital archive and the AIATSIS library. Although the recording quality is variable, potentially introducing a confounding factor, all recordings were made in the field, allowing the results from this experiment to be generalized to other underresourced language situations. Materials are not further publicly available due to usage agreements that respect Indigenous Intellectual Property; this work was done under agreements which permit work on the recordings to further Yan-nhangu and other Indigenous language work.

% \todo{acknowledge FTG0010 grant from the ELDP project somewhere when paper is not anonymous}

%TODO - data origin. Ask Claire. Include the following: when and where were the elicitations run, who transcribed them, software used. \\

Since the Yan-nhangu data originated from various different elicitation sessions, the data was preprocessed to ensure consistency. Incomplete transcriptions and audio containing English words were excluded, along with segments with no transcriptions, leaving about 156 minutes of training data. Punctuation was removed, except for apostrophes, which denote glottal stops in Yan-nhangu orthography.

The original transcriptions were all made in Yan-nhangu orthography. The phonemic transcription of a Yan-nhangu word is easily predictable from the orthographic transcription and a script was written to generate transcriptions in IPA for all Yan-nhangu annotations used during training phonemic models.

\subsection{Tokenization and Acoustic Models}
Yan-nhangu is customarily written with the same orthography used by the other Yol\textipa{N}u languages of Arnhem Land. It includes a combination of Latin letters (including \textit{y}, \textit{l}, \textit{k}, \textit{m}, and \textit{n}), modified or accented letters (including \textit{ä} and \textit{\textsubbar{t}}), digraphs (\textit{nh}, \textit{th}), and characters not otherwise found in the English segmental system (\textit{\textipa{N}}, \textit{'}). There are 25 consonants and 6 vowels. Consonant and vowel charts are given below in Table~\ref{tab:consonants} and Table~\ref{tab:vowels} respectively in IPA, with orthography in parenthesis where different. Note that for all figures in section 3.2, capitalized tokens in the IPA models represent apical consonants (e.g. ``\textit{N}" stands for \textsubbridge{n}) or long vowels (e.g. ``\textit{A}" stands for a\textlengthmark) to emphasize the one-to-one mapping between Yan-nhangu phone and phonemic token. Additionally, whitespaces are represented by underscores.

Two identical models were trained on the same data with two contrasting tokenization methods. In the first class of models, the token was defined as the grapheme, splitting digraphs like \textit{ny} /\textltailn/ into \textit{n} and \textit{y}. The orthography is the same as the Yol\textipa{N}u Matha orthography defined in \cite{yolngu-matha-zorc}. In the second class of models, the token was defined as the phoneme, such that orthographic digraphs representing one phone like \textit{ny} /\textltailn/ are one token: \textit{\textltailn}. The analysis of Yan-nhangu phonology is based on the most detailed existing documentation of Yan-nhangu phonology \cite{YanNhanju2006}.

\begin{table}[ht]
\begin{tabular}{lllllll}
\hline
p & t & {\textrtailt} (\textsubbar{t}) & \textsubbridge{t} (th) & c (tj) & k & {\textglotstop} (') \\
b & d & {\textrtaild} (\textsubbar{d}) & \textsubbridge{d} (dh) & {\textObardotlessj} (dj) & g & \\
m & n & {\textrtailn} (\textsubbar{n}) & \textsubbridge{n} (nh) & {\textltailn} (ny) & \textipa{N} & \\
 & l & {\textrtaill} (\textsubbar{l}) \\
 & r (rr) & {\textturnr} (r) \\ 
 w & & & & j (y) \\ \hline
\end{tabular}
\centering
\caption{Consonant inventory of Yan-nhangu}
\label{tab:consonants}

\end{table}
\begin{table}[ht]
\begin{tabular}{lllllll}
\cline{1-3} \cline{5-7}
i &  & u &  & {i\textlengthmark} (e) &  & {u\textlengthmark} (o) \\ 
 & a &  &  &  & {a\textlengthmark} (ä) &  \\ \cline{1-3} \cline{5-7} 
\end{tabular}
\centering
\caption{Vowel inventory of Yan-nhangu}
\label{tab:vowels}
\end{table}

Neural models trained with self-supervised pretraining on large multilingual text or speech datasets have shown great promise for under-resourced languages, particularly when data from related languages is incorporated into the training process and followed by fine-tuning on the target language \cite{Nowakowski_2023}. To this end, we use the open-source Wav2Vec2-BERT 2.0 model \cite{chung2021w2vbertcombiningcontrastivelearning} accessed from HuggingFace, which has 600 million parameters and has been pre-trained in a self-supervised manner on Facebook's multilingual corpus. Our training dataset consists of up to 156 minutes of speech, and we optimize the model using Connectionist Temporal Classification (CTC) loss with an 80/20 train-validation split. Training is performed in Jupyter notebooks on one RTX-8000 GPU. \footnote{Code will be uploaded upon acceptance} Preliminary experiments with data augmentation and attention dropout did not yield performance improvements. Each model is trained for 16 epochs using a linear learning rate scheduler, an initial learning rate of 1e-5, and early stopping to prevent overfitting. Training takes less than two hours per model.

\begin{figure}
    \centering
    \includegraphics[width=0.8\linewidth]{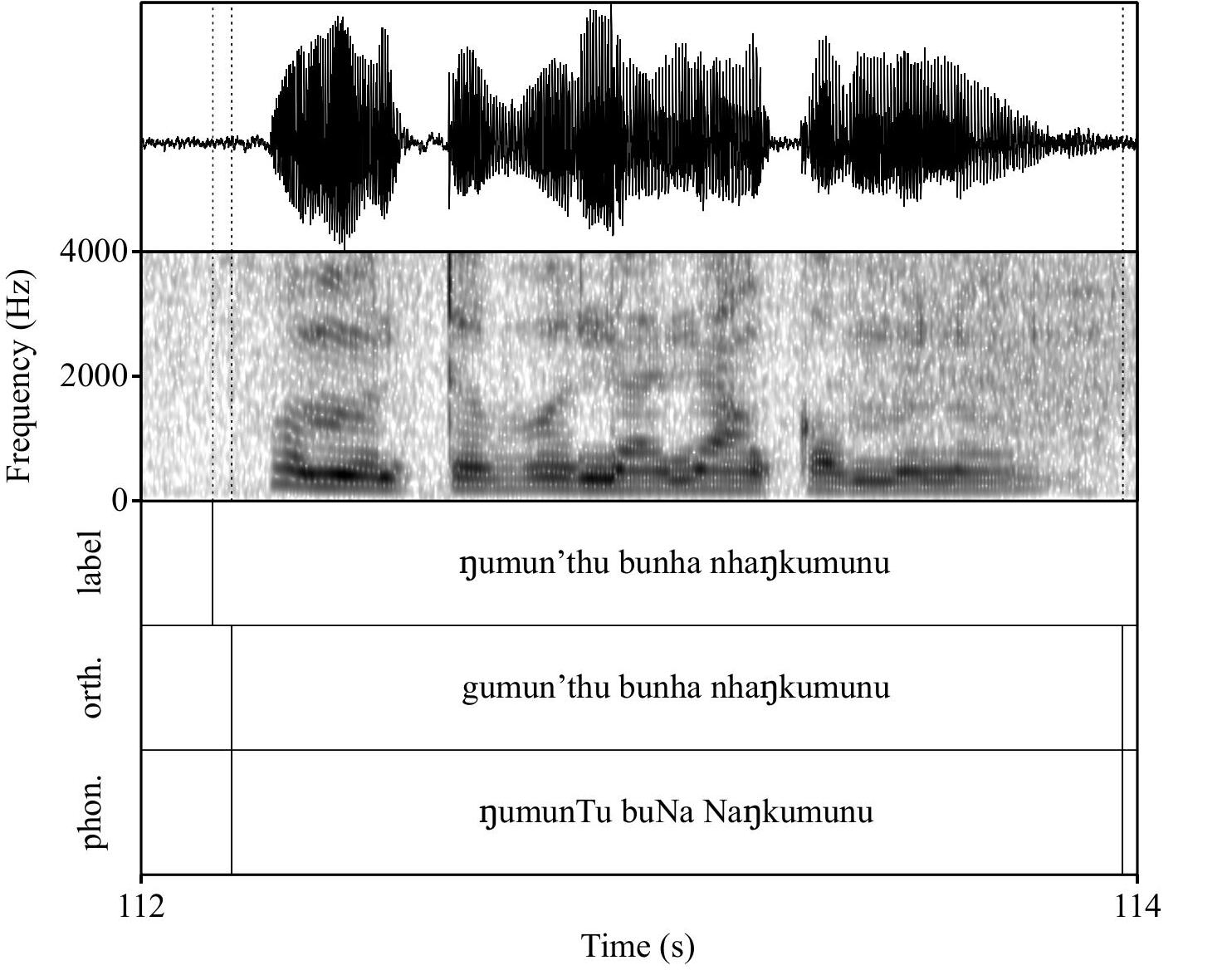}
    \caption{Waveform, spectrograph, and annotations for a sample testpoint.}
    \label{fig:sample-testpoint}
\end{figure}

Unlike traditional model evaluation, we do not create a threeway train-validation-test split. The purpose of this paper is twofold: to examine whether phonemic tokenization improves ASR model performance, and to examine whether ASR speeds up the low-resource language documentation pipeline. For this reason, we use the model with the lowest Character Error Rate for each amount of training data to compare the effects of tokenization. We use the phonemic and orthographic models with the lowest Character Error Rate to transcribe held-out validation data (Figure~\ref{fig:sample-testpoint}). To assess ASR’s feasibility in a language documentation pipeline, the last author hand-corrected four minutes of ASR output, comparing it to manual transcription from scratch (Section~\ref{sec:evaluation}).

\subsection{Evaluation}
\label{sec:evaluation}

Word Error Rate (WER) and Character Error Rate (CER) were used as evaluation metrics as defined by $\frac{S + D + I}{N}$ for Substitutions, Deletions, and Insertions for N items (characters or words) \cite{evaluate}. We select models with the lowest CER for transcription. Since the model has more parameters than training examples, training and validation loss help identify convergence issues between datasets and confirm that performance differences are not due to learning instability, but they do not necessarily reflect real-world performance. We prioritize CER when selecting top-performing models because the high frequency of single-word labels makes space-token prediction a common weakness, which distorts WER. Models with orthographic and phonemic transcription were trained on 10, 30, 60, 90, 120, and the full 156 minutes of training data, comparing WER and CER across different amounts of training data. A qualitative analysis of the automatic transcription errors most common across models was also performed and reported on in section~\ref{errors}.

Following previous ASR error analysis research \cite{ERRATTAHI201832}, we look at the Levenshtein distance between a held-out testset and the best orthographic and phonemic ASR models' transcriptions. Levenshtein distance was calculated using the \texttt{Levenshtein} package in Python and all figures were made using \texttt{matplotlib} \cite{Hunter:2007}. Figures show the 20 most frequent deletions, insertions, and substitutions for the phonemic and orthographic models with the lowest validation CER. 

Finally, the last author  manually corrected four minutes of automatically transcribed Yan-nhangu speech, such as Figure~\ref{fig:sample-testpoint}. Qualitative interpretations about the errors encountered were documented and the time save between correcting automatic transcription and manually transcribing from scratch are discussed.

\section{Results}

\subsection{WER and CER Across Models}
\begin{figure}
    \centering
    \begin{subfigure}[t]{0.23\textwidth}
        \centering
        \includegraphics[width=\linewidth]{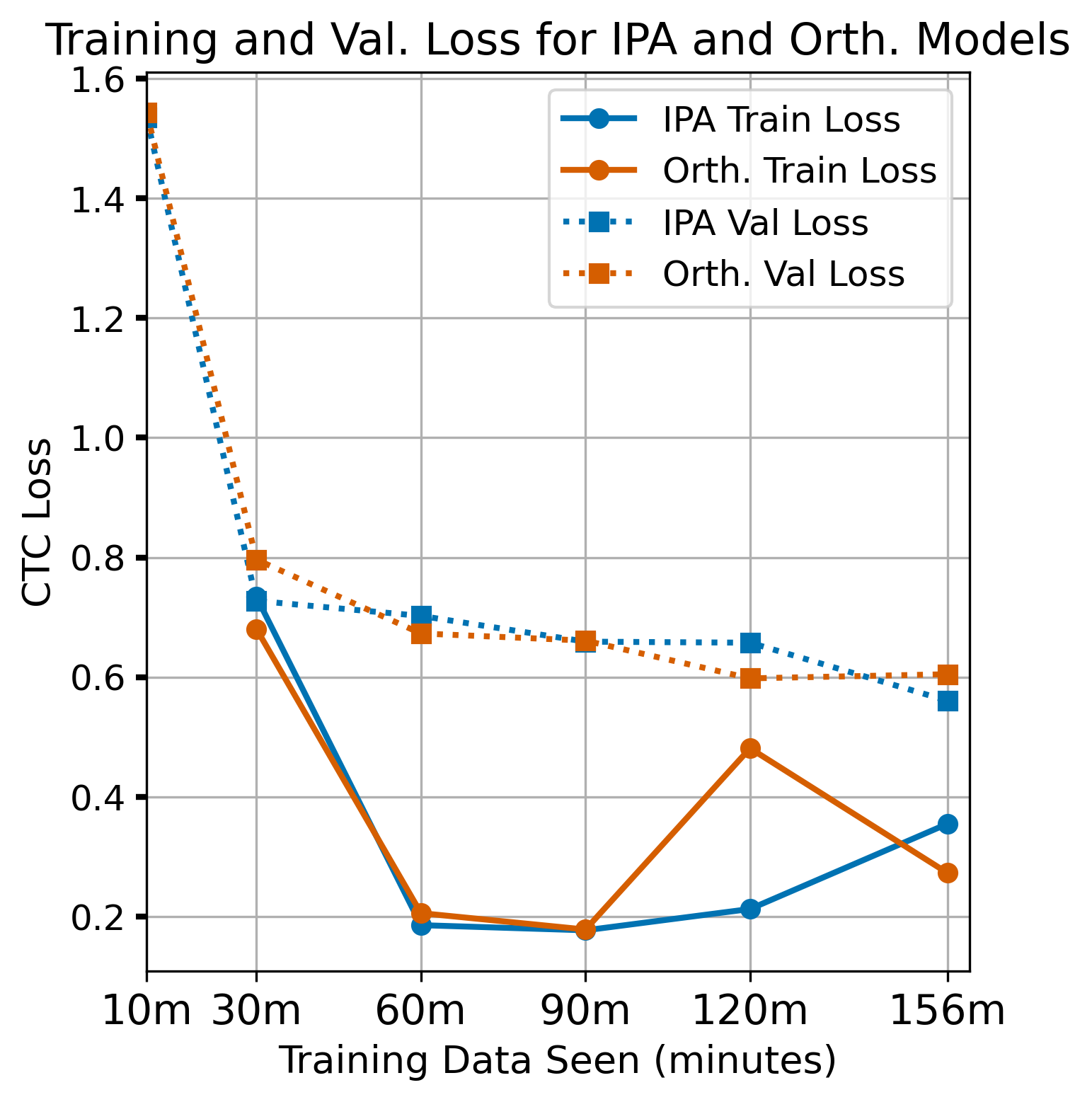} 
        \caption{Training and Validation loss for IPA and Orth. Models}
        \label{fig:losses}
    \end{subfigure}
    \hfill
    \begin{subfigure}[t]{0.23\textwidth}
        \centering
        \includegraphics[width=\linewidth]{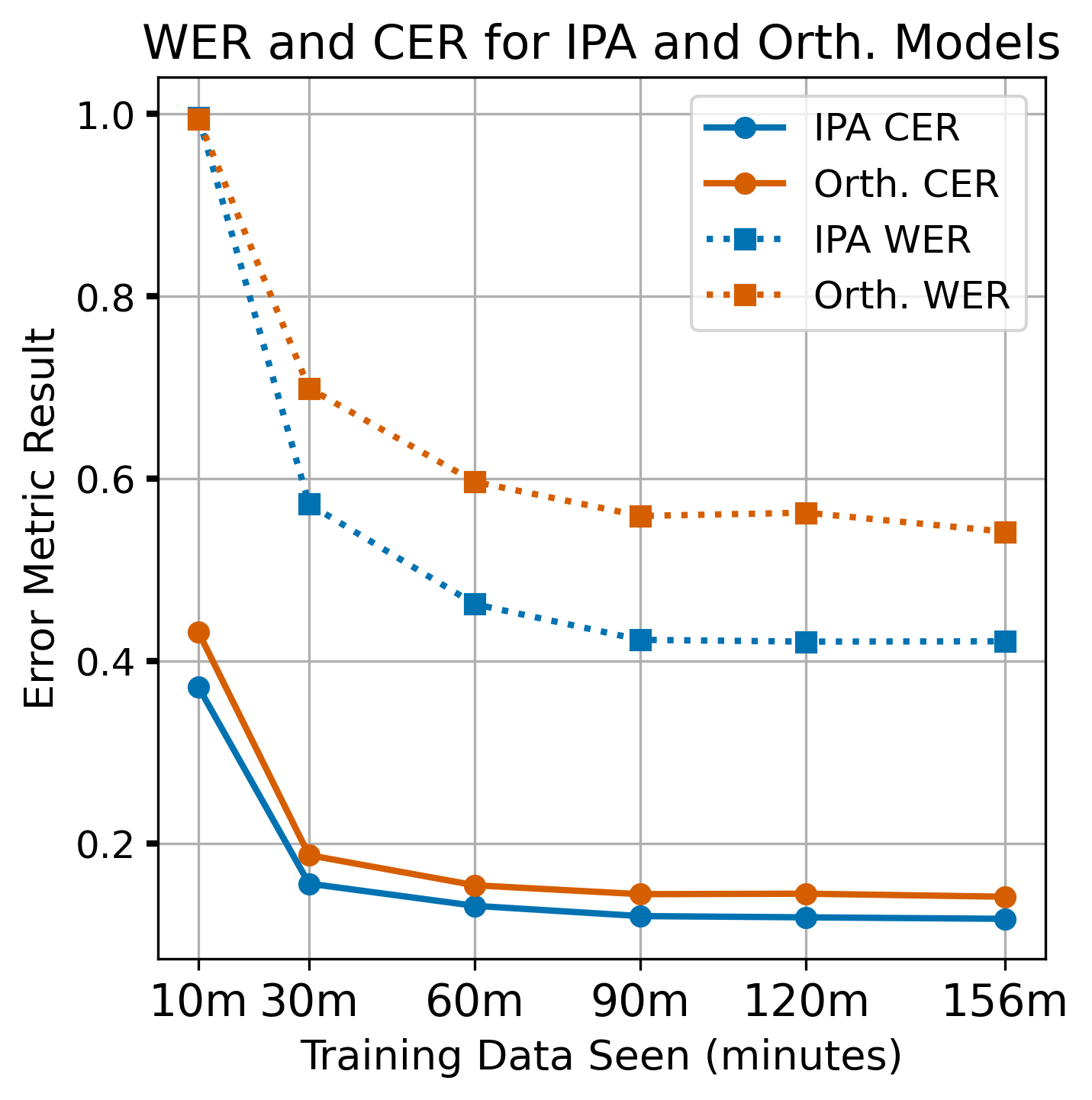} 
        \caption{CER and WER metrics for IPA and Orth. Models}
        \label{fig:cerwer}
    \end{subfigure}
    \caption{Model Comparisons by WER, CER, and loss}
\end{figure}

Our results show that phonetic tokenization improves low-resource ASR performance. As seen in Figure~\ref{fig:losses}, both models exhibit similar training and validation losses, demonstrating that they converge similarly within their respective tokenization paradigms. This indicates that performance differences are not due to underfitting or overfitting between training sets. Figure~\ref{fig:cerwer} shows a consistent gap in WER and CER on the test set, with the phonetic model outperforming the orthographic model across all training set sizes. This suggests that phonetic tokenization provides a more linguistically transparent representation of the language, reducing ambiguity and facilitating more effective generalization, particularly in the low-resource setting.

\subsection{Levenshtein Distance Analysis of Errors}

Analyzing the Levenshtein distance between manual annotations and the best orthographic and phonemic ASR outputs allows a more detailed investigation into model differences. The total number of errors made on the validation set are described in Table~\ref{tab:distance_summary}. Although the phonemic model shows lower overall Levenshtein distance from a manually transcribed reference than the orthographic model, substitutions seem to be more frequent for the phonemic model. 

\begin{table}[]
\begin{tabular}{l|llll}
Model Type & Deletions & Insertions & Substitutions & Total \\ \hline
Phonemic & 433 & 454 & 547 & 1434 \\
Orthographic & 624 & 592 & 438 & 1654
\end{tabular}
    \caption{Summary of Levenshtein distance between human annotated and ASR transcription by error type}
    \label{tab:distance_summary}
\end{table}

\begin{figure*}[h!]
    \centering
    \begin{subfigure}[t]{0.3\textwidth}
        \centering
        \includegraphics[width=\linewidth]{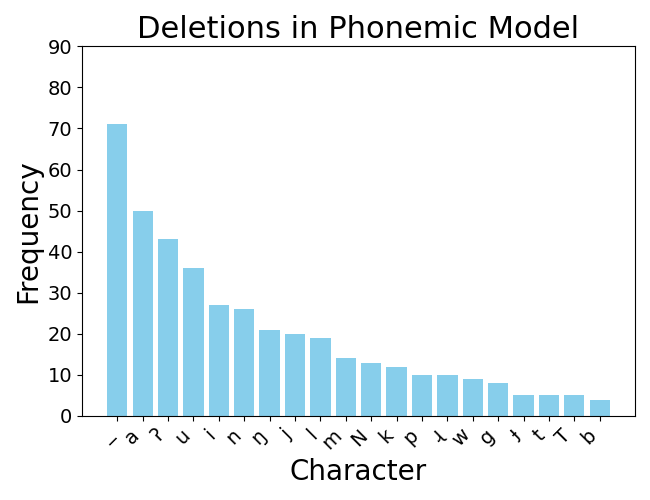} 
        \caption{Deletions for phonemic tokenization}
        \label{fig:ipa_deletions}
    \end{subfigure}
    \hfill
    \begin{subfigure}[t]{0.3\textwidth}
        \centering
        \includegraphics[width=\textwidth]{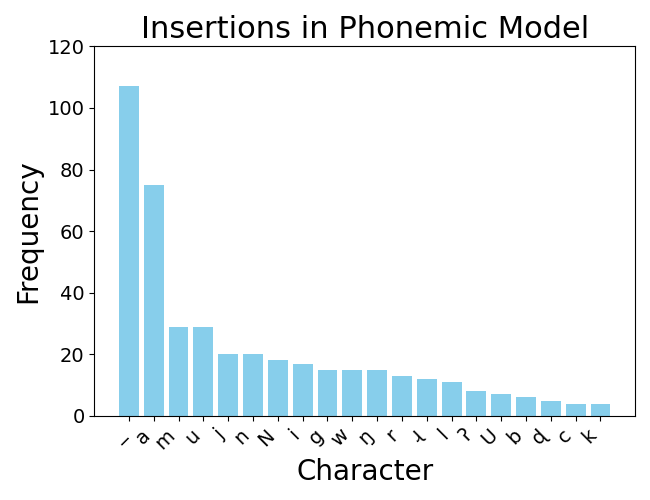} 
        \caption{Insertions for phonemic tokenization}
        \label{fig:ipa_insertions}
    \end{subfigure}
    \hfill
    \begin{subfigure}[t]{0.3\textwidth}
        \centering
        \includegraphics[width=\textwidth]{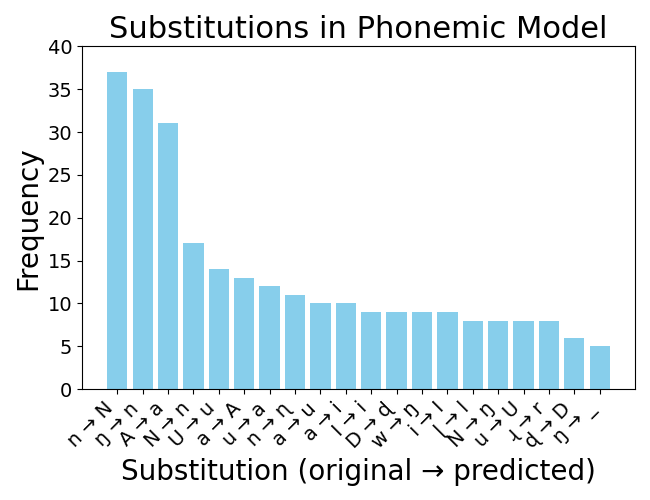} 
        \caption{Substitutions for phonemic tokenization}
        \label{fig:ipa_substitutions}
    \end{subfigure}
    \vspace{4pt}
    \begin{subfigure}[t]{0.3\textwidth}
        \centering
        \includegraphics[width=\linewidth]{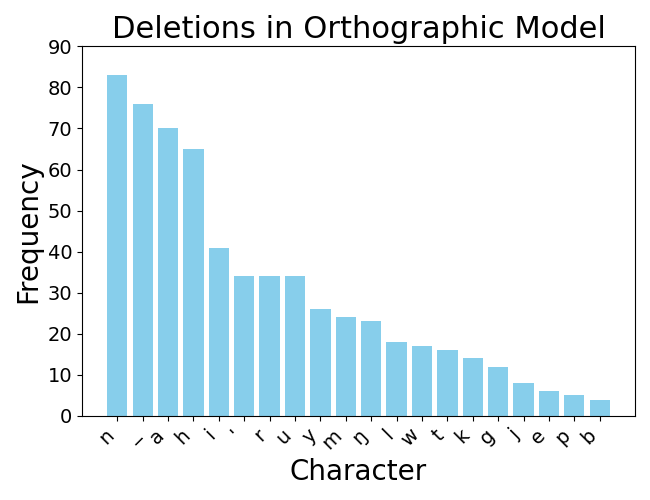} 
        \caption{Deletions for orthographic tokenization}
        \label{fig:ort_deletions}
    \end{subfigure}
    \hfill
    \begin{subfigure}[t]{0.3\textwidth}
        \centering
        \includegraphics[width=\textwidth]{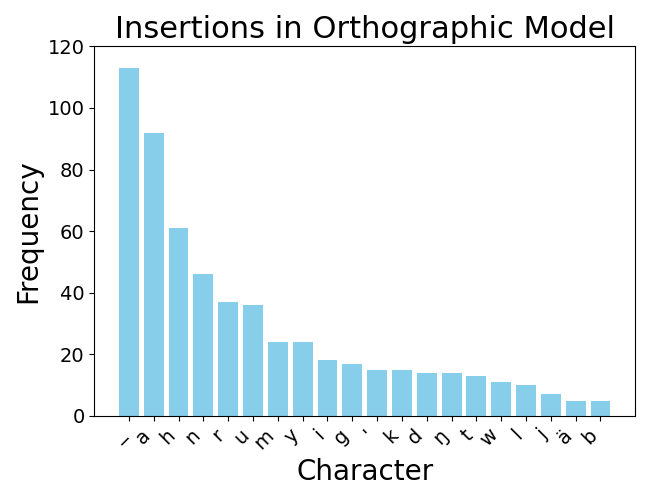} 
        \caption{Insertions for orthographic tokenization}
        \label{fig:ort_insertions}
    \end{subfigure}
    \hfill
    \begin{subfigure}[t]{0.3\textwidth}
        \centering
        \includegraphics[width=\textwidth]{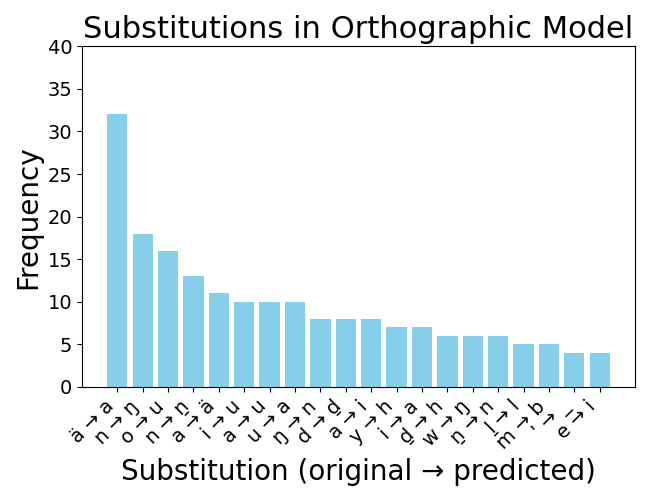} 
        \caption{Substitutions for orthographic tokenization}
        \label{fig:ort_substitutions}
    \end{subfigure}
\caption{Counts for deletions, insertions, and substitutions for the best ASR models using phonemic and orthographic tokenization.}
\label{fig:big_levenshtein}
\end{figure*}

The barcharts in Figure~\ref{fig:ipa_deletions} and Figure~\ref{fig:ort_deletions} show the frequency of each deleted character. For both orthographic and phonemic models, spaces and short vowels are among the most commonly deleted characters. This is consistent with spaces between words being unclear during rapid speech, and with vowel elision that occurs in Yan-nhangu, especially with \textit{a}. Interestingly, the orthographic model frequently deletes tokens \textit{n} and \textit{h}, both of which are present in multiple digraphs. Across both models, sonorants (nasals, liquids, and vowels) are deleted much more frequently than stops. 

Figure~\ref{fig:ipa_insertions} and Figure~\ref{fig:ort_insertions}
show the most frequently inserted characters, which mirror the most frequently deleted characters across both models. As such, whitespaces and \textit{a} are the most frequent insertions, with \textit{h} being the third most commonly inserted character for the orthographic model. Like with deletions, sonorants are inserted more frequently than stops.

Lastly, Figure~\ref{fig:ipa_substitutions} and Figure~\ref{fig:ort_substitutions}  show the most common token substitutions for the phonemic and orthographic models. This is the only setting in which we see a greater rate of errors for the phonemic model due to phonemic substitutions appearing as insertions or deletions for the orthographic model (e.g. \textit{n} $\rightarrow$ \textit{N}, corresponding to \textit{n} $\rightarrow$ \textit{nh}: in the orthographic model, this appears as an insertion). Another instance of ambiguous phones showing up as substitutions of phonologically similar phones include long vowels being substituted for their short counterparts, which occurs with an approximately equal and high frequency in both models. However, a stark asymmetry in substitution direction occurs with \textit{n} and \textit{\textipa{N}}, where \textit{\textipa{N}} $\rightarrow$ \textit{n} is the second most common substitution in the phonemic model, while \textit{n} $\rightarrow$ \textit{\textipa{N}} is the second most common substitution in the orthographic model. We theorize the cause of this discrepancy in the discussion.

\subsection{Further Comments on Errors}\label{errors}
The last author, who is very familiar with (but not fluent in) Yan-nhangu, manually corrected 4 minutes of automatically transcribed Yan-nhangu speech. It took 20 minutes to review each transcribed utterance to check for errors, to correct the errors, and make a note of the type of error. At five minutes per minute of transcript, this is equal to the fastest unassisted transcription rate (numerous factors lead to faster or slower transcription rates, including familiarity with the language, number of speakers, speech rate of participants, or complexity of the subject matter). Prior to using the ASR technology, it would take the last author approx.\ 15 minutes to transcribe 1 minute of Yan-nhangu, meaning that the introduction of ASR technology in the transcription pipeline resulted in a three times speedup.

In the stretch of recording considered, errors fell into three (quantitative) categories. The first were grapheme substitutions and word break errors. For example, ASR ran together the tense/aspect particles \emph{mana} `continuative' and \emph{gurrku} `future'; written as separate words in the Yan-nhangu orthography but frequently pronounced as a single unit. The word \emph{\underline{d}iltji} `bush' was transcribed as \emph{diltji}. The second set of ``errors'' involved items where ASR correctly rendered the material in the recording, but the transcription did not adhere to Yan-nhangu orthographic norms. For example, 2 nasals were pronounced as syllabic, and the preceding vowel was not transcribed; this correctly represents the speech on the recording but still needs to be changed to conform to Yan-nhangu orthography. The final set of 3 cases were where, to the transcriber's ear, both the ASR word and the word in the previously created transcript were possible representations of the recorded speech. The original transcripts were made in 2004--2008 and discussed with native speakers of Yan-nhangu, but there are several near homophone words that were, in context, also semantically plausible.
%\textcolor{orange}{They have found ...}

\section{Discussion and Conclusions}

This paper investigates whether phonologically informed tokenization improves ASR accuracy for the low-resource Australian language Yan-nhangu. Results show that phone-level tokenization improves WER and CER over a grapheme-level baseline for models trained on at least 30 minutes of data and improves CER across all models. Models approach peak performance with approx. 90 minutes of training data, with the orthographic model showing more room for improvement. These findings align with prior research demonstrating the benefits of linguistically informed tokenization in low-resource NLP \cite{atuhurra2024}.

Additionally, we find that changing the tokenization scheme only results in meaningful differences in substitutions, insertions, and deletions if a speech sound is represented differently by the tokenizers. In the case of our tokenization schemes, this specifically refers to speech sounds that are represented as one token phonemically but two orthographically. We theorize the discrepancy in substitution direction relating to the velar and alveolar nasal is due to the presence of \textit{n} in the digraphs ``nh" /\textsubbridge{n}/ and ``ny" /\textltailn/ for the orthographic model. This means that when the orthographic model encounters \textsubbridge{n} or \textltailn, it must represent what is one sound with two tokens. Due to its presence in digraphs, the probability of  \textit{n} is correlated to the probability of \textit{h} and \textit{y}. When the orthographic model needs to classify a velar nasal \textit{\textipa{N}}, the likelihoods of \textit{h} and \textit{y} decrease due to the former representing an apical consonant and the latter only being used for palatals, consequently decreasing the likelihood of \textit{n} as well, resulting in more frequent substitutions of \textit{\textipa{N}} for \textit{n} for the orthographic model. Note that this effect does not occur in the phonemic model since it does not contain any digraphs. Further research is needed to expand upon these preliminary findings on the effects of digraphs on ASR accuracy.

% Transcribing speech is a crucial but time consuming step in the language documentation pipeline. Many factors influence the amount of time it takes to transcribe unannotated speech, however in this experiment, correcting high-quality transcriptions provides approximately a three time speedup over the traditional method of manually transcribing the audio. Generalizing the transcription rate of the last author, using ASR to generate a first pass transcription for Yan-nhangu before manual correction provides a speedup of 10 hours for every hour of unannotated speech. Such a speedup is essential for creating more resources for the world's languages that need it the most,

Speech transcription is essential but time-consuming in language documentation. While many factors affect transcription time, this experiment shows that correcting high-quality ASR transcriptions is about three times faster than manual transcription for someone familiar with the language. Based on the last author’s transcription rate, ASR-assisted transcription provides approximately a 10-hour speedup per hour of unannotated speech. This efficiency is crucial for developing resources for the world's most under-documented languages.

\bibliographystyle{IEEEtran}
% \bibliography{mybib,field}

\end{document}